\documentclass{article}

\usepackage{arxiv}

\usepackage[utf8]{inputenc} 
\usepackage[T1]{fontenc}    
\usepackage{hyperref}       
\usepackage{url}            
\usepackage{booktabs}       
\usepackage{amsfonts}       
\usepackage{amsmath}        
\usepackage{nicefrac}       
\usepackage{microtype}      
\usepackage{cleveref}       
\usepackage{graphicx}
\usepackage{natbib}
\usepackage{doi}
\usepackage{placeins}

\title{OceanMoE: Structured Conditional Sparse Computation for Long-Horizon Multivariate Ocean Forecasting}

\author{
Yishun Zhu\textsuperscript{1,*} \quad Jian Wang\textsuperscript{2,*}\\
\small
\textsuperscript{1}Hangzhou Institute for Advanced Study, University of the Chinese Academy of Sciences\\
Zhejiang, China\\
\textsuperscript{2}Computer Network Information Center, Chinese Academy of Sciences\\
Beijing, China\\
\textsuperscript{*}Equal contribution
}

\renewcommand{\shorttitle}{OceanMoE}

\hypersetup{
colorlinks=false,
hidelinks,
pdftitle={OceanMoE: Structured Conditional Sparse Computation for Long-Horizon Multivariate Ocean Forecasting},
pdfsubject={Multivariate ocean forecasting},
pdfauthor={Yishun Zhu, Jian Wang},
pdfkeywords={Multivariate ocean forecasting, Mixture-of-Experts, Conditional computation, Long-horizon prediction},
}

\begin{document}
\maketitle

\begin{abstract}
	Multivariate ocean forecasting must exploit shared evolution in a coupled
	ocean system while adapting to the heterogeneous statistical and dynamical
	characteristics of different prediction variables and locations.
	Fully shared models may lack the flexibility to handle this heterogeneity,
	whereas fully independent models discard the common ocean context shared
	across variables.
	The key question is how to retain shared context in a unified model while
	allowing computation to specialize according to the prediction target and
	local state.
	We propose OceanMoE, a structured conditional sparse
	Mixture-of-Experts framework that combines sharing and specialization for
	multivariate ocean forecasting.
	OceanMoE fuses cross-variable information to construct target-specific local
	representations and uses them to perform content-conditioned sparse routing
	at each spatial location, with the number of active experts adapted to router
	confidence.
	In the decoder, routing is augmented with a learned geographic bias
	parameterized by spherical-harmonic spatial bases, while shared residual and
	seasonal pathways provide common cross-variable and month-dependent context.
	Experiments on long-horizon autoregressive ORAS5 forecasting show that
	OceanMoE lowers aggregate forecasting error in both evaluated settings and
	maintains lower geometric-mean normalized RMSE than the corresponding baselines over most
	later rollout months.
	Routing analyses further show that expert allocation varies with prediction
	targets and spatial locations.
These results support structured conditional computation as a modeling
strategy for balancing shared ocean context with adaptive specialization.
\end{abstract}


\section{Introduction}
\label{sec:introduction}

Data-driven forecasting is becoming an important approach to modeling complex
Earth systems. Recent neural forecasting systems trained from reanalysis data
can efficiently predict many interacting atmospheric and oceanic fields across
timescales ranging from medium-range weather to multi-year ocean variability
\citep{lam2023graphcast,guo2025orca,cui2025wenhai}. Multivariate modeling is
particularly relevant to the ocean: temperature, salinity, velocity, sea
surface height, and surface fluxes are coupled through large-scale circulation,
air--sea exchange, and the evolving ocean state. A unified model can therefore
exploit these couplings by learning a
common representation of the evolving ocean context.

Shared physical context, however, does not make ocean variables homogeneous
prediction tasks. They differ in spatial structure, temporal scale, variation
magnitude, and forecasting difficulty, motivating variable-aware
representations in Earth-system models
\citep{nguyen2023climax,guo2025orca}. Moreover, the evolution of a given
variable depends on both its local state and its geographic setting. A shared
backbone promotes information exchange by reusing the same parameterized
computation across prediction targets and spatial locations, but it does not
explicitly allocate different computation to targets and local states with
different needs. At the other extreme, training an independent model for each
variable increases specialization but fragments information that should remain
shared within the coupled system.

These two extremes expose the missing capability in multivariate ocean
forecasting: a model should determine what context remains shared and where
computation changes with the prediction conditions. Mixture-of-Experts (MoE)
models provide conditional computation by routing each representation through
a sparse subset of expert parameters, and have commonly been used to increase
model capacity without a proportional increase in computation per example
\citep{shazeer2017moe,fedus2022switch}. Multi-task MoEs further coordinate
common and task-specific parameters through a shared expert pool and
task-dependent gates \citep{ma2018mmoe}. Sparse capacity or task-level routing
alone, however, does not fully express the sharing--specialization structure of
ocean fields. The routing conditions must also reflect the prediction target,
local content, and geographic setting, while explicit shared pathways preserve
information that should not be partitioned among experts.

Following this principle, we propose \emph{OceanMoE}, a structured conditional
sparse MoE framework for multivariate ocean forecasting. For each prediction
target, OceanMoE fuses cross-variable features under target conditioning to
form target-specific local representations. The encoder then selects sparse
experts at each spatial location according to the local content state, making
the expert combination depend on the current prediction needs rather than a
predefined variable grouping. In the decoder, a learned geographic routing
bias parameterized by spherical-harmonic spatial bases introduces stable
spatial structure into expert selection, while the cumulative router confidence
determines how many experts are active at each position. Shared residual and
seasonal pathways carry common cross-variable and month-dependent context
alongside the specialized expert computation.

We evaluate OceanMoE through 60-month autoregressive forecasts on the ORAS5
ocean reanalysis \citep{zuo2019oras5},
covering a standard six-variable task and an extended ten-variable task. The
two settings compare OceanMoE with a reproduced ORCA-DL model and an
ORCA-DL-Expanded baseline with matched input and output spaces, respectively.
OceanMoE lowers aggregate forecasting error in both settings, improves
performance across most evaluated regions, and maintains lower geometric-mean normalized RMSE
than the corresponding baselines over most later rollout months. Routing
diagnostics further show that expert allocation changes with prediction targets
and spatial locations, consistent with condition-dependent computation.

The main contributions of this work are summarized as follows:
\begin{itemize}
	\item We formulate multivariate ocean forecasting as a balance between
	shared context and conditionally specialized computation, shifting the
	design question from whether to share to what to share and where to
	specialize.
		\item We introduce OceanMoE, which combines target-specific
		representations, content-conditioned per-position sparse routing, a learned
		geographic routing bias, an adaptive expert count, and shared context
		pathways to realize target- and location-dependent computation within a
		unified model.
		\item We evaluate both forecasting accuracy and routing behavior in
		long-horizon multivariate ORAS5 forecasts, showing reductions in aggregate
		error, improvements across most regions and most later rollout months, and
		target- and location-dependent routing behavior.
\end{itemize}

\section{Related Work}
\label{sec:related-work}

\subsection{Data-Driven Earth-System Forecasting}

Data-driven Earth-system forecasting learns state evolution from reanalysis
data and jointly models multiple interacting geophysical fields within a
unified network. Pangu-Weather organizes multilevel atmospheric variables into
a three-dimensional representation and forecasts them with an Earth-specific
Transformer, whereas GraphCast propagates local and global information through
a graph neural network on a multiscale spherical mesh
\citep{bi2023pangu,lam2023graphcast}. ClimaX further uses variable-specific
encoding and aggregation to accommodate different variable sets and downstream
tasks across datasets \citep{nguyen2023climax}. Together, these studies show
that shared multivariate representations can exploit coupling across Earth
system variables, and that geometry and variable organization provide useful
structural priors.

This line of work primarily addresses how coupled fields are represented and
evolved in a unified network. A complementary question is whether the
parameterized computation should also adapt to the current prediction target
and local state. OceanMoE studies this question by extending multivariate
sharing from common representations to target- and location-conditioned
computation.

\subsection{Data-Driven Ocean Forecasting}

Data-driven ocean forecasting now spans timescales from short-range,
eddy-resolving forecasts to multi-year global prediction. XiHe uses a
hierarchical Transformer with land--ocean masking and ocean-specific blocks to
model local structures and global teleconnections \citep{wang2024xihe}.
WenHai incorporates air--sea flux calculations and ocean dynamical priors into
a high-resolution forecasting system for temperature, salinity, velocity, and
sea-surface states \citep{cui2025wenhai}. At longer timescales, ORCA-DL adopts
an encoder--fusion--decoder architecture to predict three-dimensional
multivariate ocean states with atmospheric conditioning
\citep{guo2025orca}. These models demonstrate that unified data-driven
frameworks can learn coupled ocean evolution across different spatial
resolutions and forecast horizons.

The main advances along this direction concern joint ocean representation,
spatial structure, and long-horizon evolution. OceanMoE studies a complementary
axis within the unified forecasting setting: common ocean context is retained
through shared pathways, while specialized computation varies with the
prediction target and local spatial state. It therefore focuses on how sharing
and specialization are organized during joint prediction, rather than
replacing ocean-specific representations or coupling mechanisms.

\subsection{Conditional Computation and Mixture-of-Experts}

Mixture-of-Experts (MoE) models provide a general framework for conditional
computation by using a learned router to select different parameter pathways
for different inputs. Sparsely gated MoEs activate only a subset of experts,
partially decoupling model capacity from computation per example
\citep{shazeer2017moe}. GShard and Switch Transformer scale sparse experts to
large language models, while V-MoE applies token-level routing to vision
Transformers \citep{lepikhin2021gshard,fedus2022switch,riquelme2021vmoe}.
These studies establish input-conditioned sparse parameter selection as a
practical computational paradigm.

Recent work has relaxed the fixed-top-$K$ assumption by selecting experts
until their cumulative routing probability exceeds a confidence threshold,
so that different tokens can use different numbers of experts
\citep{huang-etal-2024-harder}. OceanMoE adopts this confidence-based
selection principle for spatial ocean tokens, while additionally conditioning
the router on target-specific content and geographic structure.

MoE architectures have also been used to coordinate shared and task-specific
processing in multi-task learning. Multi-gate Mixture-of-Experts (MMoE) shares
an expert pool across tasks while learning a separate gate for each task, so
that expert reuse can vary with task identity \citep{ma2018mmoe}. Spatial field
forecasting introduces a finer conditional structure: different locations of
the same target can have different local states and computational needs.
Multivariate ocean forecasting therefore calls for routing that combines
target conditioning with per-position content and geographic context.

OceanMoE connects these lines of research by constructing target-specific
representations, routing expert computation from per-position content, and
introducing geographic context into decoder routing. In parallel, shared
residual and seasonal pathways retain common cross-variable and
month-dependent context. The role of MoE in this work is thus not capacity
scaling alone, but structured sharing and specialization within a unified ocean
forecasting model.

\section{Method}
\label{sec:method}

\subsection{Preliminaries}

\noindent\textbf{Multivariate ocean forecasting.} Let
$\mathcal{X}_{t}=\{\mathbf{x}_{t}^{(v)}\}_{v=1}^{V}$ denote the multivariate
ocean state at month $t$, where
$\mathbf{x}_{t}^{(v)}\in\mathbb{R}^{C_v\times H\times W}$ contains variable
$v$ over $C_v$ vertical levels. Let
$\mathcal{Q}\subseteq\{1,\ldots,V\}$ be the set of prediction targets and
$\mathcal{A}_{t}$ the auxiliary atmospheric surface fields. Given the current
ocean state, atmospheric context, and calendar month, a forecasting model
estimates
\begin{equation}
    \widehat{\mathcal{X}}^{\mathcal{Q}}_{t+1:t+T}
    = F_{\theta}(\mathcal{X}_{t},\mathcal{A}_{t},t).
    \label{eq:forecast_problem}
\end{equation}
Inputs and targets are normalized with month-dependent statistics estimated
from the training set, and predictions are transformed back to physical units
for evaluation.

For target $v$, the prediction head directly produces the normalized field
$\widehat{\mathbf{y}}_{t+\delta}^{(v)}$ at lead $\delta$. Land cells are masked
before patch embedding and excluded from the objective. Let
$\mathcal{I}_{\mathrm{ocean}}$ denote the fixed global ocean grid and define
the latitude-area weights once for the full grid as
\begin{equation}
    w_i=\cos(\varphi_i),
    \qquad
    \widetilde{w}_i
    =\frac{w_i}
    {\frac{1}{|\mathcal{I}_{\mathrm{ocean}}|}
     \sum_{j\in\mathcal{I}_{\mathrm{ocean}}}w_j}.
    \label{eq:global_area_weight}
\end{equation}
The same normalized weights are used for every sample, variable, month, and
region; they are not recomputed from sample-specific valid cells. We compute a
latitude-area-weighted root mean squared error separately for each target,
\begin{equation}
    \mathcal{L}_{v}^{\mathrm{pred}}
    = \sqrt{
      \frac{\sum_{c,i}M_{v,c,i}\widetilde{w}_i
      (\widehat{y}_{v,c,i}-y_{v,c,i})^2}
      {\sum_{c,i}M_{v,c,i}\widetilde{w}_i}
      +\epsilon},
    \label{eq:prediction_loss}
\end{equation}
where $M_{v,c,i}$ is the ocean mask, $\epsilon=10^{-12}$, and the mean of
$\widetilde{w}_i$ over $\mathcal{I}_{\mathrm{ocean}}$ is one. This RMSE is
computed in the normalized training space, with no additional variable-scale
normalization. No depth weighting is applied, so the $C_v$ levels of a
three-dimensional variable contribute equally within that variable.

For a training batch, let $\mathcal{B}_v$ be the samples with a valid label for
variable $v$ and let $n_v=|\mathcal{B}_v|$. The per-variable reductions above
are aggregated over the variables present in the batch as
\begin{equation}
    \mathcal{L}_{\mathrm{pred}}
    = \frac{\sum_{v\in\mathcal{Q}_{\mathrm{batch}}}
      n_v\mathcal{L}_{v}^{\mathrm{pred}}}
      {\sum_{v\in\mathcal{Q}_{\mathrm{batch}}}n_v},
    \label{eq:aggregate_prediction_loss}
\end{equation}
where $\mathcal{Q}_{\mathrm{batch}}$ contains the variables with at least one
valid label. Thus, when all variables are present, the variable losses are
equally weighted; variables without valid labels are skipped.

The model is trained over a direct forecast window of $M$ months and extended
to longer horizons through block-autoregressive inference. Let
$t_j=t+jM$ denote the beginning of rollout block $j$. The model predicts
\begin{equation}
    \widehat{\mathcal{X}}_{t_j+1:t_j+M}^{\mathcal{Q}}
    = F_{\theta}(\widehat{\mathcal{X}}_{t_j},
      \mathcal{A}^{\mathrm{cond}}_{j},t_j),
    \qquad
    \widehat{\mathcal{X}}_{t_{j+1}}
    \leftarrow \widehat{\mathcal{X}}_{t_j+M},
    \label{eq:block_rollout}
\end{equation}
where $\mathcal{A}^{\mathrm{cond}}_{j}$ denotes the aligned zonal and
meridional atmospheric surface-stress fields for the $M$ target months in
block $j$. These fields are encoded separately from the ocean state and
supplied as prescribed conditioning for the corresponding direct forecast
block; they are not predicted by the ocean decoder. The direct window and
rollout horizon are specified in Section~\ref{sec:experiments}. The rollout
state contains the complete model ocean-variable contract; variables without
a continuous evaluation target remain input/context channels rather than
reported forecast targets.

\noindent\textbf{Parameter-sharing formulations.} Two limiting formulations
clarify how computation can be organized across prediction targets. A fully
shared forecaster uses one parameter set for every target,
\begin{equation}
    \widehat{\mathbf{x}}_{t+\delta}^{(q)}
    = F_{\theta}(\mathcal{X}_{t},\mathcal{A}_{t},q,\delta),
    \label{eq:fully_shared_forecaster}
\end{equation}
where the target identity $q$ conditions a common mapping $F_{\theta}$. At the
other endpoint, a target-specific formulation assigns an independent parameter
set to each target,
\begin{equation}
    \widehat{\mathbf{x}}_{t+\delta}^{(q)}
    = F_{\theta_q}(\mathcal{X}_{t},\mathcal{A}_{t},\delta).
    \label{eq:target_specific_forecaster}
\end{equation}
Both formulations may access the same coupled ocean state; they differ in
whether the parameterized computation is reused across targets. We use them as
conceptual endpoints of the sharing design space rather than as an exhaustive
taxonomy of forecasting architectures.

\noindent\textbf{Sparse conditional computation.} Given a local feature
$\mathbf{h}_{i}\in\mathbb{R}^{d}$, a router produces logits
$\boldsymbol{\ell}_{i}\in\mathbb{R}^{E}$ over $E$ experts. The routing
probability of expert $e$ is
\begin{equation}
    p_{i,e}
    = \frac{\exp(\ell_{i,e})}
    {\sum_{j=1}^{E}\exp(\ell_{i,j})},
    \label{eq:router_probability}
\end{equation}
For an active expert set $\mathcal{S}_{i}$, a sparse MoE computes
\citep{shazeer2017moe}
\begin{equation}
    \operatorname{MoE}(\mathbf{h}_{i})
    = \sum_{e\in\mathcal{S}_{i}}
      \widetilde{p}_{i,e}f_e(\mathbf{h}_{i}),
    \qquad
    \widetilde{p}_{i,e}
    = \frac{p_{i,e}}{\sum_{j\in\mathcal{S}_{i}}p_{i,j}},
    \label{eq:sparse_moe}
\end{equation}
so only the selected experts are evaluated and their probabilities are
renormalized within the active set.

\subsection{Our Method}

\begin{figure*}[t]
    \centering
    \includegraphics[width=0.98\textwidth]{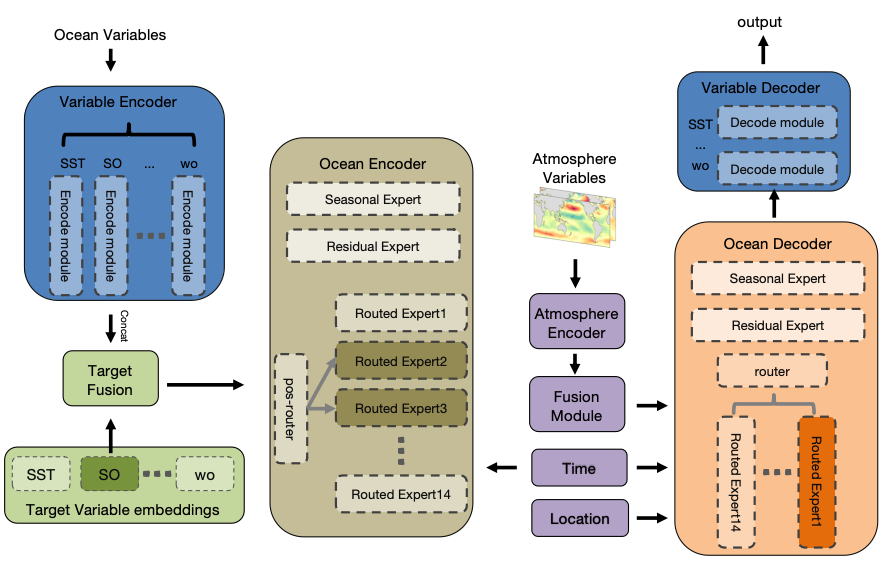}
    \caption{Overview of OceanMoE. Variable-wise ocean encoders feed the
    target-specific fusion module, and the figure's ``Atmosphere Variables''
    branch denotes the two prescribed surface stresses $(\tau_u,\tau_v)$.
    In both Ocean Encoder and Ocean Decoder, the boxes labeled ``Seasonal
    Expert'' and ``Residual Expert'' correspond to the shared seasonal and
    residual pathways in Eq.~\eqref{eq:shared_fusion}, while ``Routed Expert''
    boxes denote the adaptively selected private experts. The Time branch
    supplies calendar-month conditioning, and the Location branch supplies the
    spherical-harmonic geographic bias to the decoder router. Variable-wise
    decoders produce a six-month forecast block, whose final state is fed into
    the next block during the 60-month rollout.}
    \label{fig:architecture}
\end{figure*}

OceanMoE is designed around three linked decisions. First, it learns which
source-variable information is relevant to each prediction target at each
location. Second, it provides parameter-shared pathways for context that need
not be mediated solely by expert selection. Third, it allocates specialized computation
according to both the target-conditioned local state and geographic structure,
including the amount of expert capacity required at each position. The
resulting execution order is
\emph{variable-wise encoding} $\rightarrow$ \emph{target-specific fusion}
$\rightarrow$ \emph{encoder routing and shared fusion}
$\rightarrow$ \emph{decoder routing with a geographic bias}
$\rightarrow$ \emph{target-wise prediction}. This decomposition turns the
question of ``shared or independent'' computation into the more specific
questions of what information should be shared, where computation should
specialize, and how much specialized capacity a local state requires.

In the following, $s\in\mathcal{B}=\{\mathrm{enc},\mathrm{dec}\}$ denotes a
routed stage and $q$ denotes a prediction target; these indices specialize the
generic MoE quantities in Eqs.~\eqref{eq:router_probability}--
\eqref{eq:sparse_moe}. Figure~\ref{fig:architecture} summarizes the data
flow. Variable-specific encoders are kept separate until target-specific
fusion, while the atmospheric surface stresses are encoded by a separate
branch and remain external block conditioning rather than predicted outputs.
The routed and shared computations below operate on the resulting target-wise
states before each prediction head reconstructs its future field.

\noindent\textbf{Target-specific cross-variable representation.} Let
$\mathbf{z}_{i}^{(v)}\in\mathbb{R}^{d}$ be the encoded feature of source
variable $v$ at bottleneck position $i\in\{1,\ldots,N\}$, where $N=hw$, and
let $\mathbf{e}_{v}$ denote its learned variable embedding. For target
$q\in\mathcal{Q}$, we construct a target query and source keys,
\begin{align}
    \mathbf{q}_{i}^{(q)}
    &= W_{Q}^{c}\operatorname{LN}(\mathbf{z}_{i}^{(q)})
      + W_{Q}^{a}\operatorname{LN}(\mathbf{e}_{q}), \\
    \mathbf{k}_{i}^{(v)}
    &= W_{K}^{c}\operatorname{LN}(\mathbf{z}_{i}^{(v)})
      + W_{K}^{a}\operatorname{LN}(\mathbf{e}_{v}).
    \label{eq:target_query_key}
\end{align}
Their patch-local compatibility is
\begin{equation}
    s_{q,v,i}
    = \frac{\langle\mathbf{q}_{i}^{(q)},\mathbf{k}_{i}^{(v)}\rangle}
    {\sqrt{d}}
    + b_{\mathrm{self}}\mathbb{I}[q=v],
    \label{eq:target_compatibility}
\end{equation}
where $b_{\mathrm{self}}$ is a learnable self-bias that provides a preference
for the source variable corresponding to the target. We normalize
$\mathbf{s}_{q,i}=[s_{q,1,i},\ldots,s_{q,V,i}]$ over source variables with a
sparse transformation,
\begin{equation}
    \boldsymbol{\alpha}_{q,i}
    = \operatorname{entmax}_{1.5}(\mathbf{s}_{q,i}),
    \label{eq:target_weights}
\end{equation}
where $\operatorname{entmax}_{1.5}$ produces a normalized, potentially sparse
source-weight distribution \citep{peters2019sparse}. We then obtain the
target-specific local state
\begin{equation}
    \mathbf{b}_{q,i}
    = \sum_{v=1}^{V}\alpha_{q,v,i}P_v\mathbf{z}_{i}^{(v)}
      + \mathbf{b}_{f}.
    \label{eq:target_fusion}
\end{equation}
Here, $P_v$ is a source-specific projection and $\mathbf{b}_f$ is a learnable
fusion bias. Target
identity therefore affects subsequent routing through the fused content state
$\mathbf{b}_{q,i}$ rather than through a target-only expert gate. The weights
$\alpha_{q,v,i}$ provide an interpretable description of how source variables
contribute to each target-specific state.

\noindent\textbf{Structured conditional routing.} At the encoder fusion
stage, the target-specific local state determines both the local expert input
and the router logits,
\begin{equation}
    \mathbf{u}^{\mathrm{enc}}_{q,i}=\mathbf{b}_{q,i},
    \qquad
    \boldsymbol{\ell}^{\mathrm{enc}}_{q,i}
    = W_{r}^{\mathrm{enc}}\operatorname{LN}(\mathbf{b}_{q,i})
      + \mathbf{c}_{r}^{\mathrm{enc}}.
    \label{eq:encoder_router}
\end{equation}
No target embedding is concatenated directly to this router. Target dependence
is already encoded in $\mathbf{b}_{q,i}$, while position-specific content
allows different locations of the same target to activate different expert
combinations. The selected experts transform
$\mathbf{u}^{\mathrm{enc}}_{q,i}$, after which their output is combined with
the shared pathways in Eq.~\eqref{eq:shared_fusion}.

At the decoder stage, the fused encoder outputs are projected back to the
decoder grid and combined with the target-specific shared context to form
$\mathbf{d}_{q,i}\in\mathbb{R}^{d}$. No additional cross-variable fusion is
performed after this point. The decoder state is passed to the content router,
the geographic router, and the prediction head in that order. The content
component of the decoder router is
\begin{equation}
    \mathbf{u}^{\mathrm{dec}}_{q,i}=\mathbf{d}_{q,i},
    \qquad
    \boldsymbol{\ell}^{\mathrm{content}}_{q,i}
    = W_{r}^{\mathrm{dec}}\operatorname{LN}(\mathbf{d}_{q,i})
      + \mathbf{c}_{r}^{\mathrm{dec}}.
    \label{eq:decoder_content_router}
\end{equation}
To represent stable global spatial structure, we evaluate real spherical
harmonics up to degree $L=4$ at the latitude--longitude coordinate of position
$i$. Excluding the constant component gives
$\boldsymbol{\phi}_{i}\in\mathbb{R}^{(L+1)^2-1}=\mathbb{R}^{24}$. A learnable geographic
router maps this basis to expert logits,
\begin{align}
    \widetilde{\mathbf{g}}_i
    &= W_{g,2}\operatorname{SiLU}
       (W_{g,1}\boldsymbol{\phi}_i+\mathbf{b}_{g,1})+\mathbf{b}_{g,2}, \\
    \mathbf{g}_i
    &= \sigma(s_g)
       \frac{\widetilde{\mathbf{g}}_i
       -\operatorname{mean}_{e}(\widetilde{g}_{i,e})}
       {\operatorname{RMS}(\widetilde{\mathbf{g}}_i)+\epsilon}.
    \label{eq:geo_router}
\end{align}
where $s_g$ is a learnable scalar controlling the magnitude of the geographic
contribution.
The final decoder logits are
\begin{equation}
    \boldsymbol{\ell}^{\mathrm{dec}}_{q,i}
    = \boldsymbol{\ell}^{\mathrm{content}}_{q,i}+\mathbf{g}_i.
    \label{eq:decoder_router}
\end{equation}
The geographic bias is shared across targets and depends only on location,
whereas target dependence enters through $\mathbf{d}_{q,i}$. It therefore acts
as an additive spatial prior rather than a fixed assignment of experts to
geographic regions. Decoder routing is conditioned jointly on the current
target-specific state and this learned global spatial prior.

After either router produces probabilities
$\mathbf{p}^{s}_{q,i}$, OceanMoE converts the router confidence into an
adaptive expert capacity following the confidence-based dynamic-selection
principle of Huang et al.~\citep{huang-etal-2024-harder}. Let
$\pi^{s}_{q,i}$ order experts by decreasing probability and define the
smallest cumulative-probability set
\begin{align}
    K^{*,s}_{q,i}
    &= \min\left\{k\in\{1,\ldots,E\}:
      \sum_{j=1}^{k}p^{s}_{q,i,\pi^{s}_{q,i}(j)}\ge p_0\right\}, \\
    K^{s}_{q,i}
    &= \min\left(K_{\max},
       \max\left(K_{\min},K^{*,s}_{q,i}\right)\right).
    \label{eq:dynamic_k}
\end{align}
The active set $\mathcal{S}^{s}_{q,i}$ contains the first
$K^{s}_{q,i}$ experts under $\pi^{s}_{q,i}$. Their probabilities are
renormalized as in Eq.~\eqref{eq:sparse_moe}, and sparse dispatch evaluates
only the selected token--expert pairs. Concentrated router distributions
therefore use fewer experts, whereas diffuse distributions can use more,
subject to the explicit $K_{\min}$ and $K_{\max}$ bounds. Router logits are
perturbed during training and are deterministic at inference; the perturbation
schedule and selection hyperparameters are given in the experimental setup.

\noindent\textbf{Shared context pathways.} Sparse routing allocates
conditional computation, but shared information should not have to pass only
through the selected private experts. We therefore construct a target-aware
global context using projections shared across targets. We first pool each
target-specific bottleneck representation and modulate it with an affine
transformation derived from the target embedding,
\begin{align}
    \overline{\mathbf{b}}_{q}
    &= \operatorname{LN}\left(\frac{1}{N}
       \sum_{i=1}^{N}\mathbf{b}_{q,i}\right), \\
    [\boldsymbol{\gamma}_{q},\boldsymbol{\beta}_{q}]
    &= W_{\mathrm{AdaLN}}\operatorname{LN}(\mathbf{e}_{q}), \\
    \mathbf{c}_{q}
    &= (1+\boldsymbol{\gamma}_{q})\odot\overline{\mathbf{b}}_{q}
       +\boldsymbol{\beta}_{q}.
    \label{eq:attribute_context}
\end{align}
The global context $\mathbf{c}_{q}$ is used by the shared pathways and is not
concatenated to the per-position encoder router in
Eq.~\eqref{eq:encoder_router}; the encoder router remains driven by the local
target-specific state.

For each routed stage $s\in\mathcal{B}=\{\mathrm{enc},\mathrm{dec}\}$, let
$\mathbf{u}^{s}_{q,i}$ be its local input, $\mathbf{m}^{s}_{q,i}$ its expert
output, and $\mathbf{e}^{\mathrm{mon}}_t$ the calendar-month embedding. The
seasonal pathway is
\begin{equation}
    \mathbf{s}^{s}_{q,i}
    = P^{s}_{\mathrm{sea}}\operatorname{LN}
      (\mathbf{u}^{s}_{q,i}+\mathbf{c}_{q}+\mathbf{e}^{\mathrm{mon}}_t),
    \label{eq:seasonal_path}
\end{equation}
and the residual pathway is
\begin{equation}
    \mathbf{r}^{s}_{q,i}
    = P^{s}_{\mathrm{res}}
      \left[
        \operatorname{LN}(\mathbf{u}^{s}_{q,i});
        \operatorname{sg}(\operatorname{LN}(\mathbf{m}^{s}_{q,i}));
        \operatorname{sg}(\operatorname{LN}(\mathbf{s}^{s}_{q,i}))
      \right],
    \label{eq:residual_path}
\end{equation}
where $[\cdot;\cdot]$ denotes concatenation and $\operatorname{sg}$ denotes
stop-gradient. The stage output is
\begin{equation}
    \mathbf{h}^{s}_{q,i}
    = \mathbf{m}^{s}_{q,i}
      + \eta^{s}_{\mathrm{res}}\mathbf{r}^{s}_{q,i}
      + \eta^{s}_{\mathrm{sea}}\mathbf{s}^{s}_{q,i}.
    \label{eq:shared_fusion}
\end{equation}
The projections in these pathways are shared across targets, while
$\mathbf{u}^{s}_{q,i}$ and $\mathbf{c}_{q}$ preserve target-dependent state.
The stop-gradient operators prevent the residual pathway from changing the
expert and seasonal representations through its auxiliary inputs; the two
paths can still receive gradients through their own direct contributions in
Eq.~\eqref{eq:shared_fusion}. The learnable scales
$\eta^{s}_{\mathrm{res}}$ and $\eta^{s}_{\mathrm{sea}}$ are initialized at
zero and learned jointly with the routed pathway, so training starts from the
routed computation and learns how much shared context to add.

\noindent\textbf{Routing differentiation objective.} For each routed stage
$s\in\mathcal{B}$, we average the router probabilities over positions for each
target,
\begin{equation}
    \overline{\mathbf{p}}^{s}_{q}
    = \frac{1}{N}\sum_{i=1}^{N}\mathbf{p}^{s}_{q,i}.
    \label{eq:mean_router_distribution}
\end{equation}
Their pairwise Jensen--Shannon divergence is
\begin{equation}
    \mathcal{D}_{\mathrm{route}}
    = \frac{1}{|\mathcal{B}|\binom{|\mathcal{Q}|}{2}}
      \sum_{s\in\mathcal{B}}\sum_{q<q'}
      \operatorname{JS}
      (\overline{\mathbf{p}}^{s}_{q}\,\|\,
       \overline{\mathbf{p}}^{s}_{q'}).
    \label{eq:routing_diversity}
\end{equation}
The complete training objective is
\begin{equation}
    \mathcal{L}
    = \mathcal{L}_{\mathrm{pred}}
      - \lambda_{\mathrm{div}}\mathcal{D}_{\mathrm{route}}.
    \label{eq:total_loss}
\end{equation}
Because the diversity term enters with a negative coefficient, the objective
encourages target-level routing distributions to differentiate while the
prediction loss maintains forecasting accuracy. During autoregressive inference,
the target-specific representations and routing decisions are recomputed
whenever the model receives the updated state from the preceding rollout
block.

\section{Experiments}
\label{sec:experiments}

\subsection{Datasets}

We use three data sources with distinct roles. Historical CMIP6 simulations
provide the training archive, SODA2 provides reanalysis data for
training-time validation, and ORAS5 provides the fixed long-horizon
forecasting benchmark. All evaluated models use the common preprocessing and
forecast protocol described below.

\noindent\textbf{CMIP6 training archive.} The training data consist of monthly
historical simulations from 20 models participating in CMIP6
\citep{eyring2016cmip6}. The local archive spans January 1850 to December
2014. The training loader uses January 2010 as an exclusive endpoint, giving
an effective training period from January 1850 through December 2009.
The 14-variable models read 14 ocean variables and two atmospheric surface
stress variables from each CMIP6 source. The reproduced six-variable baseline
uses the same climate models and training months but selects the standard six
ocean prediction variables together with the same two surface-stress inputs.
This construction keeps the underlying climate-model archive fixed while
allowing the two forecasting contracts to use their declared output spaces.
The six-variable comparison in the results uses the 14-variable OceanMoE
checkpoint scored on the six complete ORAS5 targets, whereas reproduced
ORCA-DL is trained with the six-variable input/output contract; this is
therefore a cross-contract comparison rather than a matched parameter-space
ablation.

\noindent\textbf{SODA2 validation archive.} SODA2
\citep{carton2008soda} supplies monthly salinity, potential temperature,
sea-surface temperature, zonal and meridional velocity, sea-surface height,
and zonal and meridional surface stress from 1871 through 2010; sea-surface
salinity is available through 1979 in the local archive. Targets unavailable
in a given reanalysis sample are excluded from that sample's objective.

\noindent\textbf{ORAS5 forecasting benchmark.} ORAS5
\citep{zuo2019oras5} provides a complete initialization in December 2020 and
monthly targets from January 2021 through December 2025 for all ten evaluated
variables. Every model starts from the same December 2020 ocean state and
produces a 60-month autoregressive forecast using the configurations and
checkpoints listed in Table~\ref{tab:model_architectures}.
The aligned surface stresses for each future block are
supplied as prescribed external conditioning and are passed unchanged to every
compared model.

\noindent\textbf{Variable inventory.} The standard task evaluates potential
temperature, salinity, zonal and meridional velocity, sea-surface temperature,
and sea-surface height. The extended task additionally evaluates downward
surface heat flux, mixed-layer thickness, sea-surface salinity, and water flux
into seawater. Four further ocean variables participate in the 14-variable
model but are excluded from 60-month error metrics because a complete ORAS5
target sequence is unavailable. The complete model inventory is therefore the
six standard targets \texttt{so}, \texttt{thetao}, \texttt{tos}, \texttt{uo},
\texttt{vo}, and \texttt{zos}; the four added evaluation targets \texttt{hfds},
\texttt{mlotst}, \texttt{sos}, and \texttt{wfo}; and the four model-only
ocean variables \texttt{rsntds}, \texttt{sob}, \texttt{tob}, and \texttt{wo}.
The atmospheric inputs are the single-level stresses \texttt{tauu} and
\texttt{tauv}. Three-dimensional ocean fields are represented at 16 vertical
levels, whereas the remaining fields are single-level.

\subsection{Model Architectures and Experimental Setup}

\noindent\textbf{Reproduced ORCA-DL.} ORCA-DL consists of variable-wise ocean
encoders, an atmosphere encoder, a fusion module, and variable-wise ocean
decoders \citep{guo2025orca}. The ocean variables and surface stresses are
encoded separately, the fusion module integrates their latent features under
lead-time conditioning, and the decoders reconstruct the future field of each
ocean variable with encoder--decoder skip connections. Its maximum direct
forecast interval is six months; longer forecasts are generated through
block-autoregressive rollout. Our reproduced model follows this six-variable
dense architecture and contains 540.5M parameters.

\noindent\textbf{ORCA-DL-Expanded.} ORCA-DL-Expanded preserves the dense
encoder--fusion--decoder organization of ORCA-DL while expanding its ocean
encoders, fusion inputs, and prediction heads to the same 14-variable model
space used by OceanMoE. It contains 883.4M parameters and provides the
ten-variable comparison with a matched ocean input and output contract.

\noindent\textbf{OceanMoE.} OceanMoE first constructs a target-specific
representation by fusing cross-variable features. Its encoder routes each
spatial position according to the resulting local content state, while its
decoder augments content-dependent routing with a learned geographic bias
parameterized by spherical-harmonic bases. Shared residual and seasonal
pathways carry common context alongside the sparsely selected experts, as
detailed in Section~\ref{sec:method}. OceanMoE contains 936.2M parameters. Of
these, 37.2M belong to private experts, 5.5M to routing and target
conditioning, and 13.3M to shared pathways; the remaining parameters belong
to the variable-wise encoder--decoder backbone. Sparse dispatch evaluates only
the selected token--expert pairs.

\begin{table}[htbp]
    \centering
    \small
    \caption{Architectural roles and evaluated checkpoints.}
    \label{tab:model_architectures}
    \begin{tabular}{lcccc}
        \toprule
        Model & Computation & Model variables & Parameters & Checkpoint \\
        \midrule
        Reproduced ORCA-DL & Dense shared fusion & 6 & 540.5M & 16,000 \\
        ORCA-DL-Expanded & Dense shared fusion & 14 & 883.4M & 28,000 \\
        OceanMoE & Conditional sparse fusion & 14 & 936.2M & 12,000 \\
        \bottomrule
    \end{tabular}
\end{table}

\noindent\textbf{Preprocessing.} All fields are bilinearly remapped to a
$128\times360$ latitude--longitude grid with $1^{\circ}$ spacing, covering
$63.5^{\circ}$S--$63.5^{\circ}$N and $0.5^{\circ}$E--$359.5^{\circ}$E.
Three-dimensional variables are vertically interpolated to 10, 15, 30, 50,
75, 100, 125, 150, 200, 250, 300, 400, 500, 600, 800, and 1000~m. Each field
is normalized with month-dependent mean and standard-deviation statistics
estimated from the shared training archive. Predictions and targets are
returned to physical units before evaluation.

Land cells are masked before patch embedding and excluded from evaluation.

\noindent\textbf{Training configuration.} All three models use AdamW, cosine
learning-rate decay, a warmup ratio of 0.1, weight decay of 0.1, random seed
1, the same training archive, effective batch size, BF16 precision,
learning-rate setting ($2\times10^{-4}$), and total training budget. The
parameter-count differences arise from the model architectures rather than
from different data or optimization budgets; the six- and fourteen-variable
contracts differ only in their declared input/output spaces. Evaluation uses
checkpoint 16,000 for reproduced ORCA-DL, 28,000 for ORCA-DL-Expanded, and
12,000 for OceanMoE, as listed in Table~\ref{tab:model_architectures}.

OceanMoE uses $E=14$ experts and dynamic top-$p$ routing with cumulative
probability threshold $p_0=0.8$, $K_{\min}=1$, and $K_{\max}=14$. The
geographic router uses real spherical harmonics through degree $L=4$, a
hidden width of 64, an initial scale of 0.5, and a 500-step warmup. Noisy
routing is used only during training, with standard
deviation initialized at 0.05 and capped at 0.5; inference is deterministic.
The routing differentiation coefficient is $\lambda_{\mathrm{div}}=10^{-3}$
in Eq.~\eqref{eq:total_loss}.

For the component-removal sensitivity analysis, all five models are evaluated
over the same 60-month ORAS5 window and all 70 target layers, using the same
latitude-area weighting and physical-space RMSE as the main evaluation. The
four variants retain the complete forecasting contract while
removing one design choice or replacing dynamic routing with fixed top-2
selection.

\noindent\textbf{Rollout protocol.} Each model directly predicts a six-month
block and passes the final predicted state to the next block, producing ten
blocks and 60 forecast months in total. Within every pairwise comparison, the
models use the same initialization and target months. Each
block also receives the aligned auxiliary atmospheric surface fields as
conditioning; the ocean model does not autoregressively predict these fields.

\noindent\textbf{Evaluation metrics.} Land cells are excluded with the ocean
mask. For variable $v$, level $c$, forecast month $m$, and region
$\mathcal{R}$, the latitude-area-weighted RMSE is
\begin{equation}
    \operatorname{RMSE}_{v,c,m}^{\mathcal{R}}
    = \sqrt{
      \frac{\sum_{i\in\mathcal{R}}M_{v,c,i}\widetilde{w}_i
      (\widehat{y}_{v,c,m,i}-y_{v,c,m,i})^2}
      {\sum_{i\in\mathcal{R}}M_{v,c,i}\widetilde{w}_i}}.
    \label{eq:eval_rmse}
\end{equation}
Here, $M_{v,c,i}$ is the ocean mask and $\widetilde{w}_i$ is the fixed global
normalized latitude-area weight defined in Eq.~\eqref{eq:global_area_weight}.
The weights are not renormalized after restricting the calculation to a
sample, variable, month, or region.
We first average the physical-unit RMSE in Eq.~\eqref{eq:eval_rmse} over
all 60 months, denoting this arithmetic time average by
$\overline{\operatorname{RMSE}}$. To summarize relative performance across
variables with different units and numbers of vertical levels, we use a
hierarchically weighted geometric mean of RMSE ratios, following the
cross-series relative-error aggregation principle of
\citet{kourentzes2019pooling}. Each variable receives equal total weight,
and its levels are weighted equally:
\begin{equation}
    R^{\mathcal R}
    = \exp\!\left[
      \frac{1}{|\mathcal V|}\sum_{v\in\mathcal V}\frac{1}{C_v}
      \sum_{c=1}^{C_v}\log
      \frac{\overline{\operatorname{RMSE}}_{v,c,\mathrm{OceanMoE}}^{\mathcal R}}
           {\overline{\operatorname{RMSE}}_{v,c,\mathrm{base}}^{\mathcal R}}
      \right],\qquad
    \Delta^{\mathcal R}=100(R^{\mathcal R}-1).
    \label{eq:aggregate_relative_rmse}
\end{equation}
Negative $\Delta^{\mathcal R}$ indicates lower OceanMoE error. This
aggregation measures relative performance on a multiplicative scale, avoids
mixing physical units, and prevents variables with more levels from receiving
more total weight. It also makes the model comparison invariant to a positive
normalization factor shared by both models at each variable and level. This
is an aggregate relative-error score, not a pooled physical RMSE: it allows
trade-offs across targets, so we also report variable-wise physical RMSE.
All RMSE values entering the reported geometric aggregates are positive and
finite; no clipping or additive offset is applied.

For regional comparisons, we display a geometric-mean normalized score using
the corresponding baseline's Global Ocean error as a fixed reference:
\begin{equation}
    G_r^{\mathcal R}
    = \exp\!\left[
      \frac{1}{|\mathcal V|}\sum_{v\in\mathcal V}\frac{1}{C_v}
      \sum_{c=1}^{C_v}\log
      \frac{\overline{\operatorname{RMSE}}_{v,c,r}^{\mathcal R}}
           {\overline{\operatorname{RMSE}}_{v,c,\mathrm{base}}^{\mathrm{Global}}}
      \right].
    \label{eq:regional_global_normalization}
\end{equation}
Only the Global Ocean baseline score is constrained to one. Because the
common references cancel, $G_{\mathrm{OceanMoE}}^{\mathcal R}/
G_{\mathrm{base}}^{\mathcal R}=R^{\mathcal R}$ in every region. Thus, the
reported reduction $100(1-R^{\mathcal R})$ is consistent with both the
score columns and Eq.~\eqref{eq:aggregate_relative_rmse}. The two tasks
use different baselines, so score magnitudes are compared within each task.

For the forecast-month curves, we apply the same geometric aggregation to
the RMSE at month $m$, retaining the fixed full-window global reference:
\begin{equation}
    S_{r,m}
    = \exp\!\left[
      \frac{1}{|\mathcal V|}\sum_{v\in\mathcal V}\frac{1}{C_v}
      \sum_{c=1}^{C_v}\log
      \frac{\operatorname{RMSE}_{v,c,m,r}^{\mathrm{Global}}}
           {\overline{\operatorname{RMSE}}_{v,c,\mathrm{base}}^{\mathrm{Global}}}
      \right].
    \label{eq:monthly_normalized_rmse}
\end{equation}
The curves describe month-specific relative performance. Their time averages
need not equal the full-window scores, which average physical RMSE over time
before geometric aggregation across levels and variables. In particular,
the baseline curve is not constrained to have a time mean of one. Figure
annotations use the full-window result from
Eq.~\eqref{eq:aggregate_relative_rmse}, not the mean of the plotted curves.

Regional analysis uses the Global Ocean and six fixed diagnostic boxes:
Pacific ($120^{\circ}$E--$290^{\circ}$E), Atlantic
($290^{\circ}$E--$20^{\circ}$E across the longitude seam), Indian
($20^{\circ}$E--$120^{\circ}$E), Southern Ocean
($63.5^{\circ}$S--$60^{\circ}$S), northern high-latitude ocean
($60^{\circ}$N--$63.5^{\circ}$N), and equatorial Pacific
($5^{\circ}$S--$5^{\circ}$N, $120^{\circ}$E--$290^{\circ}$E). Because the
evaluation grid ends at $63.5^{\circ}$ latitude, the two high-latitude boxes
represent narrow boundary bands rather than complete polar basins.

\noindent\textbf{Reproducibility.} We retain every denormalized prediction,
target, and error array at shape $60\times128\times360$ for each evaluated
variable and vertical level. The experiment artifacts additionally include
regional and horizon summaries and full routing tensors on the $16\times30$
grid. Array shapes, metric row counts, and routing-probability normalization
are checked programmatically before aggregation.

\subsection{Results and Analysis}

\noindent\textbf{Q1: Does OceanMoE improve aggregate long-horizon
multivariate forecasting accuracy over the two baselines?}\\[0.25em]
\textbf{A1.} Yes under both declared forecasting contracts.
OceanMoE reduces the geometric-mean relative RMSE by 4.91\% relative to the reproduced ORCA-DL
model on the standard six-variable task and by 8.34\% relative to
ORCA-DL-Expanded on the extended ten-variable task. These results establish
consistent aggregate gains under both evaluated forecasting contracts.

\noindent\textbf{Q2: How does OceanMoE perform across prediction
variables?}\\[0.25em]
\textbf{A2.} Performance varies across prediction targets.
OceanMoE lowers error for salinity, potential temperature, and sea-surface height on the six-variable task. On the ten-variable task, the largest reductions occur for sea-surface salinity, sea-surface height, and sea-surface temperature. The mixed variable-wise pattern is consistent with target-dependent specialization rather than a uniform gain across fields.

\noindent\textbf{Q2a: Is the ten-variable gain driven by one target?}\\[0.25em]
\textbf{A2a.} No. Removing sea-surface salinity still gives a 5.91\% aggregate reduction over the remaining nine variables. The gain therefore extends beyond that target, although its magnitude depends on the evaluated variable set.

\begin{table}[htbp]
    \centering
    \scriptsize
    \setlength{\tabcolsep}{2.5pt}
    \caption{Variable-wise Global Ocean RMSE in physical units. For
    three-dimensional variables, each entry is the mean of the 16 level-wise
    RMSE values. Lower values indicate better predictive accuracy.}
    \label{tab:variable_results}
    \begin{tabular}{llrrrr}
        \toprule
        & & \multicolumn{2}{c}{Six-variable task} &
        \multicolumn{2}{c}{Ten-variable task} \\
        \cmidrule(lr){3-4}\cmidrule(lr){5-6}
        Variable & Unit & ORCA-DL & OceanMoE & ORCA-DL-Expanded & OceanMoE \\
        \midrule
        \texttt{so} & psu & 0.20638 & 0.17021 & 0.17058 & 0.169834 \\
        \texttt{thetao} & $^{\circ}$C & 1.04771 & 0.92827 & 0.95943 & 0.928232 \\
        \texttt{tos} & $^{\circ}$C & 0.97971 & 1.05116 & 1.36120 & 1.051150 \\
        \texttt{uo} & m\,s$^{-1}$ & 0.08461 & 0.09599 & 0.08624 & 0.095956 \\
        \texttt{vo} & m\,s$^{-1}$ & 0.06521 & 0.06755 & 0.06593 & 0.067546 \\
        \texttt{zos} & m & 0.09324 & 0.08119 & 0.10806 & 0.081194 \\
        \texttt{hfds} & W\,m$^{-2}$ & -- & -- & 79.95500 & 83.715613 \\
        \texttt{mlotst} & m & -- & -- & 41.44452 & 36.449152 \\
        \texttt{sos} & psu & -- & -- & 0.67675 & 0.490427 \\
        \texttt{wfo} & kg\,m$^{-2}$\,s$^{-1}$ & -- & -- & $1.0824\!\times\!10^{-4}$ & 0.000107614 \\
        \bottomrule
    \end{tabular}
\end{table}

\noindent\textbf{Q3: How does the relative performance vary across
geographic regions?}\\[0.25em]
\textbf{A3.} Table~\ref{tab:regional_results} shows lower geometric-mean
normalized RMSE in six of the seven reported regions on the six-variable
task and in all seven on the ten-variable task. The six-variable task has
its largest reductions in the northern high-latitude ocean (14.03\%) and
equatorial Pacific (13.95\%), but degrades in the Southern Ocean (4.66\%).
On the ten-variable task, the largest reductions occur in the northern
high-latitude ocean (19.06\%), Southern Ocean (18.47\%), and Atlantic
(11.23\%); the equatorial Pacific improves by 2.50\%. These differences
show that relative performance depends on the target set and geographic
context.

\begin{table}[htbp]
    \centering
    \scriptsize
    \setlength{\tabcolsep}{3.2pt}
    \caption{Regional comparison using geometric-mean normalized
    RMSE $G_r^{\mathcal{R}}$ from Eq.~\eqref{eq:regional_global_normalization}.
    Each variable-level error is normalized by the corresponding baseline's
    Global Ocean 60-month mean RMSE, using the same reference for all regions
    within a task. Lower scores are better; bold marks the lower score in
    each model pair. Reduction is $100(1-G_{\mathrm{OceanMoE}}/G_{\mathrm{base}})$,
    computed before rounding; negative values indicate higher OceanMoE error.}
    \label{tab:regional_results}
    \begin{tabular}{lrrrrrr}
        \toprule
        & \multicolumn{3}{c}{Six-variable task} &
        \multicolumn{3}{c}{Ten-variable task} \\
        Region & \shortstack{Baseline\\NRMSE} &
        \shortstack{OceanMoE\\NRMSE} & \shortstack{Error\\reduction (\%)} &
        \shortstack{Baseline\\NRMSE} & \shortstack{OceanMoE\\NRMSE} &
        \shortstack{Error\\reduction (\%)} \\
        \cmidrule(lr){2-4}\cmidrule(lr){5-7}
        \midrule
        Global Ocean & 1.0000 & \textbf{0.9509} & 4.91 & 1.0000 & \textbf{0.9166} & 8.34 \\
        Pacific & 0.9730 & \textbf{0.9004} & 7.47 & 0.8988 & \textbf{0.8409} & 6.44 \\
        Atlantic & 0.9964 & \textbf{0.9701} & 2.64 & 1.1282 & \textbf{1.0014} & 11.23 \\
        Indian & 1.0283 & \textbf{1.0066} & 2.11 & 0.9775 & \textbf{0.9240} & 5.47 \\
        Southern Ocean & \textbf{0.6886} & 0.7207 & -4.66 & 0.8367 & \textbf{0.6821} & 18.47 \\
        Northern high latitude & 0.8868 & \textbf{0.7624} & 14.03 & 1.1744 & \textbf{0.9506} & 19.06 \\
        Equatorial Pacific & 1.0193 & \textbf{0.8771} & 13.95 & 0.7905 & \textbf{0.7707} & 2.50 \\
        \bottomrule
    \end{tabular}
\end{table}

\begin{figure}[htbp]
    \centering
    \includegraphics[width=\linewidth]{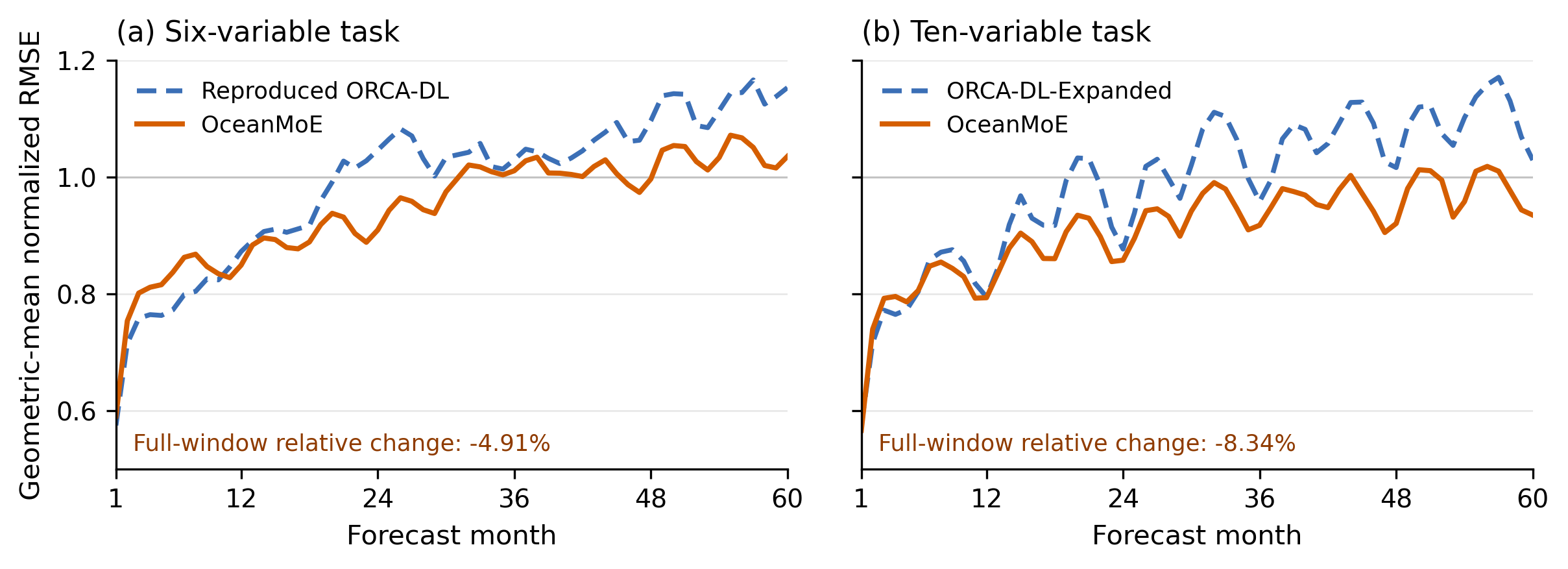}
    \caption{Monthly geometric-mean normalized RMSE $S_{r,m}$ from
    Eq.~\eqref{eq:monthly_normalized_rmse} over the 60-month
    autoregressive rollout. Panel (a) compares the six-variable reproduced
    ORCA-DL contract with OceanMoE; panel (b) compares ORCA-DL-Expanded with
    OceanMoE on the ten-variable contract. Each point corresponds to one
    forecast month; lower values indicate lower normalized
    error. For each task, both curves use the corresponding baseline's
    variable- and level-wise 60-month mean RMSE as the denominator. The plots
    show lead-time variation and are not cumulative pooled horizon scores.
    The annotations report the full-window geometric-mean relative change,
    computed after averaging each variable-level physical RMSE over 60 months;
    they are not calculated from the time means of the plotted curves.}
    \label{fig:monthly_overall_rmse}
\end{figure}

\noindent\textbf{Q4: How does the relative performance change with forecast
horizon?}\\[0.25em]
\textbf{A4.} Figure~\ref{fig:monthly_overall_rmse} shows the geometric-mean
normalized RMSE at each forecast month of the 60-month autoregressive rollout.
The curves fluctuate with lead time rather than improving monotonically, but
OceanMoE is below its corresponding baseline for most of the later rollout
months in both contracts: over months 13--60, it is lower in all 48 months
for both tasks. Early rollout months nevertheless include crossings between
the two models. This view complements the full-window aggregate results by showing
when the relative advantage appears and how stable it is across individual
lead months.

At the target level, the full-window trend is led by salinity, potential
temperature, and sea-surface height, as shown in
Table~\ref{tab:variable_results}. For sea-surface height, the 60-month mean
RMSE decreases from 0.09324 to 0.08119\,m on the six-variable task, a reduction
of approximately 12.9\%. Figure~\ref{fig:monthly_overall_rmse} complements
these full-window results by showing the month-to-month variation in
aggregate error.

\FloatBarrier

\noindent\textbf{Q5: Do the main design components improve balanced
multivariate forecasting?}\\[0.25em]
\textbf{A5.} Table~\ref{tab:component_ablation} compares the complete model
with four matched sensitivity variants on the ten-variable, full-depth
evaluation. For model $m$, we use the same geometric aggregation relative to the complete model
\begin{equation}
    S_m = \exp\!\left[\frac{1}{|\mathcal V|}
    \sum_{v\in\mathcal V}\frac{1}{C_v}
    \sum_{c=1}^{C_v}\log
    \frac{\overline{\operatorname{RMSE}}_{v,c,m}}
         {\overline{\operatorname{RMSE}}_{v,c,\mathrm{Full}}}\right],
    \label{eq:ablation_overall}
\end{equation}
so that the complete model has $S_{\mathrm{Full}}=1$. Removing the shared
context pathways, target-specific fusion, or decoder geographic routing
increases the aggregate error ratio by 1.20\%, 1.84\%, and 1.92\%,
respectively. Replacing dynamic expert selection with fixed top-2 routing has
the largest effect, increasing it by 17.69\%. Some variants improve individual
targets, but none improves the variable-balanced geometric-mean score. The result
therefore favors the complete configuration under the stated multivariate score
rather than claiming uniform gains for every target.

\begin{table}[htbp]
    \centering
    \scriptsize
    \setlength{\tabcolsep}{2.4pt}
    \caption{Ten-variable, full-depth component-removal sensitivity analysis.
    Per-variable entries are physical-unit RMSE; three-dimensional variables
    are averaged over 16 level-wise RMSE values. Overall change is
    $100(S_m-1)$, where $S_m$ is the geometric-mean ratio in
    Eq.~\eqref{eq:ablation_overall}. Lower is better; bold
    marks the lowest RMSE in each row.}
    \label{tab:component_ablation}
    \begin{tabular}{llrrrrr}
        \toprule
        Variable & Unit & Full & \shortstack{$-$ Shared\\Context} &
        \shortstack{$-$ Decoder\\GeoRouter} & \shortstack{$-$ Target\\Fusion} &
        \shortstack{Fixed\\$K=2$} \\
        \midrule
        \texttt{so} & psu & \textbf{0.169834} & 0.171304 & 0.171469 & 0.171091 & 0.246157 \\
        \texttt{thetao} & $^{\circ}$C & 0.928232 & 0.927402 & \textbf{0.923289} & 0.932451 & 1.125637 \\
        \texttt{tos} & $^{\circ}$C & 1.051150 & 1.048703 & 1.058691 & \textbf{1.040879} & 1.144868 \\
        \texttt{uo} & m\,s$^{-1}$ & 0.095956 & 0.092966 & 0.090916 & \textbf{0.090498} & 0.096402 \\
        \texttt{vo} & m\,s$^{-1}$ & 0.067546 & \textbf{0.065959} & 0.066595 & 0.066851 & 0.070754 \\
        \texttt{zos} & m & \textbf{0.081194} & 0.085898 & 0.082664 & 0.087629 & 0.118985 \\
        \texttt{hfds} & W\,m$^{-2}$ & 83.715613 & 84.555988 & \textbf{82.520360} & 83.305787 & 94.041140 \\
        \texttt{mlotst} & m & \textbf{36.449152} & 36.604106 & 36.781655 & 36.462973 & 37.823342 \\
        \texttt{sos} & psu & \textbf{0.490427} & 0.546658 & 0.626686 & 0.593008 & 0.856603 \\
        \texttt{wfo} & kg\,m$^{-2}$\,s$^{-1}$ & 0.000107614 & 0.000107717 & 0.000107876 & 0.000105714 & \textbf{0.000088479} \\
        \midrule
        Overall change & \% & \textbf{0.00} & +1.20 & +1.92 & +1.84 & +17.69 \\
        \bottomrule
    \end{tabular}
\end{table}

\begin{figure}[htbp]
    \centering
    \includegraphics[width=\linewidth]{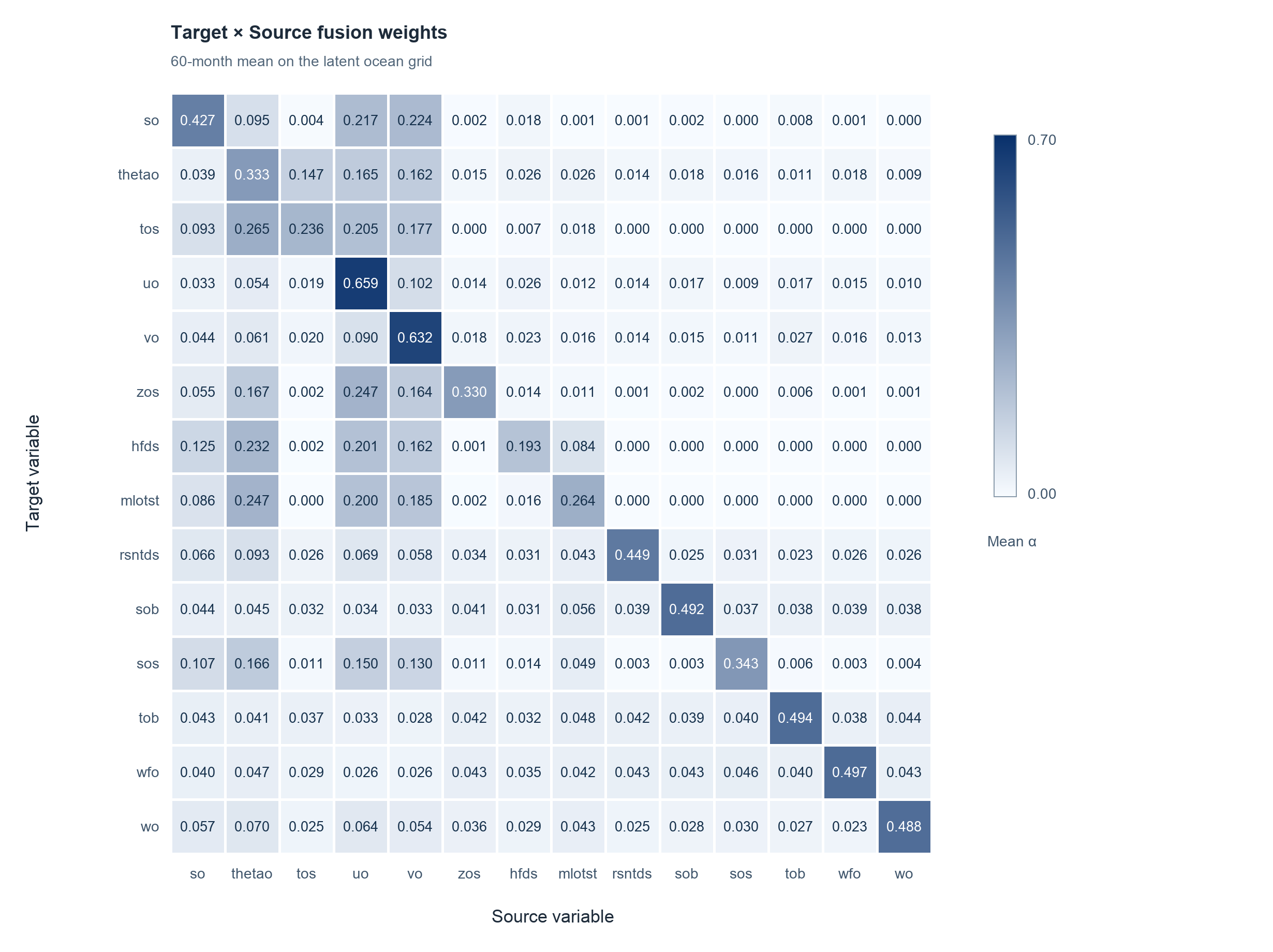}
    \caption{Target-specific selection of source features before expert
    routing in OceanMoE. Rows denote target variables and columns denote
    source variables across all 14 modeled ocean variables. Entries are the
    60-month means of $\alpha_{q,v,i}$, with latent positions weighted by
    their ocean fractions; patches with zero ocean fraction are excluded.
    Darker blue indicates a larger mean fusion weight, and each row sums to
    one before rounding. Diagonal and off-diagonal entries describe
    same-variable and cross-variable feature selection, respectively. The
    different row patterns show how shared ocean information is combined
    differently for each prediction target.}
    \label{fig:target_source_alpha}
\end{figure}

\noindent\textbf{Q6: What does the target--source heatmap reveal?}\\[0.25em]
\textbf{A6.} Figure~\ref{fig:target_source_alpha} shows how target-specific fusion combines
the encoded ocean variables before expert routing. Each row corresponds to
one prediction target, and its weights multiply the projected source features
$P_v\mathbf{z}_{i}^{(v)}$ in Eq.~\eqref{eq:target_fusion} to form that
target's local state. Diagonal entries describe the use of the target's own
variable, while off-diagonal entries describe the use of other ocean variables.

The matrix reveals distinct row patterns rather than a fixed source-variable grouping. This is physically plausible: ocean currents are linked to sea-surface-height gradients, while temperature and salinity jointly determine density and stratification; surface fluxes and mixed-layer depth also provide context for upper-ocean temperature and salinity evolution. The model reflects these relationships in its feature mixtures. Velocity targets retain large self-weights of 0.659 for \texttt{uo} and 0.632 for \texttt{vo}, whereas \texttt{tos} combines \texttt{thetao}, \texttt{tos}, \texttt{uo}, and \texttt{vo} with weights totaling 0.883. The \texttt{zos} row also emphasizes velocity and potential-temperature features. Thus, the heatmap provides a compact diagnostic of physically coherent cross-variable coupling and target-specific selection before routing. These weights support consistency with known ocean structure, but do not establish causal influence.

\noindent\textbf{Q7: Does expert allocation vary across prediction
targets?}\\[0.25em]
\textbf{A7.} Routing statistics are computed over all 14 model
targets, all 60 rollout months, and the native $16\times30$ routing grid.
Table~\ref{tab:routing_summary} reports the mean active-expert count $K$ and
the effective number of experts under normalized selected load. The
sea-surface-salinity example has a different encoder and decoder active count
from the all-target average, providing descriptive evidence that allocation is
target-dependent.

The configuration names in Table~\ref{tab:routing_summary} are compact run
labels: the first term identifies the encoder-side conditioning, the second
identifies the decoder-side conditioning, and the numeric suffix is the saved
checkpoint step. ``Geo'' denotes the decoder geographic bias, whereas
``Content'' denotes content-only routing at that stage. AdaLN+Geo-12000 is the
OceanMoE configuration used for the main results.

For reproducibility, all routing summaries use the ocean fraction of the
native $16\times30$ patches as the spatial weight and average over the 14
model targets and 60 rollout months before reporting a configuration-level
value. If $\ell_e$ is the normalized selected load of expert $e$, the
selected-effective count is $(\sum_e \ell_e^2)^{-1}$. Spatial JS is the
area-weighted Jensen--Shannon divergence between each valid patch's mean
routing distribution and the global mean distribution. The reported SH
$R^2$ is the corresponding area-weighted coefficient of determination when
the mean routing-probability map is fitted with the spherical-harmonic basis
used in Eq.~\eqref{eq:geo_router}; these diagnostics characterize the spatial
organization of routing.

\begin{table}[htbp]
    \centering
    \small
    \setlength{\tabcolsep}{4pt}
    \caption{Routing statistics averaged over all 14 model targets and 60
    forecast months. Eff. denotes the effective number of experts under
    normalized selected load.}
    \label{tab:routing_summary}
    \begin{tabular}{lrrrrrr}
        \toprule
        Configuration & Enc. $K$ & Enc. eff. & Dec. $K$ & Dec. eff. & Spatial JS & SH $R^2$ \\
        \midrule
        AdaLN + Geo-12000 & 1.76 & 3.22 & 2.15 & 6.93 & 0.0318 & 0.935 \\
        Content + Geo-12000 & 2.67 & 5.38 & 4.06 & 7.66 & 0.0451 & 0.934 \\
        Mask-aware-8000 & 2.20 & 5.01 & 2.02 & 6.44 & 0.1447 & 0.809 \\
        Content + Content-12000 & 2.56 & 4.78 & 1.58 & 5.18 & 0.0057 & 0.548 \\
        \bottomrule
    \end{tabular}
\end{table}

Both GeoRouter configurations obtain a decoder spherical-harmonic weighted
$R^2$ of approximately 0.935, indicating stable large-scale geographic
organization in their mean routing maps. The OceanMoE configuration
activates only 1.76 encoder experts and 2.15 decoder experts per position on
average, while its effective expert counts across targets, locations, and
months reach 3.22 and 6.93. Sparse local computation therefore coexists with
broader expert utilization over the rollout.

Sea-surface salinity illustrates how target selectivity and geographic
diversity can coexist. Its encoder activates $K=1.03$ experts on average,
whereas its decoder activates $K=2.63$ experts and retains a
spherical-harmonic weighted $R^2$ of 0.936. These diagnostics indicate that
expert allocation changes across targets and locations.

The target-wise summaries refine this observation. Across the four
configurations, encoder routing differs between targets by mean absolute
amounts of 0.0937--0.1033. Decoder differentiation is strongest for
Content+Geo (0.0503) and weakest for Mask-aware (0.0000), whose decoder maps
are shared across targets because target-specific decoder conditioning is
disabled. Thus, spatial structure and target specialization are separate
properties: a route can be spatially organized without being target-specific.

\begin{figure}[t]
    \centering
    \includegraphics[width=\columnwidth]{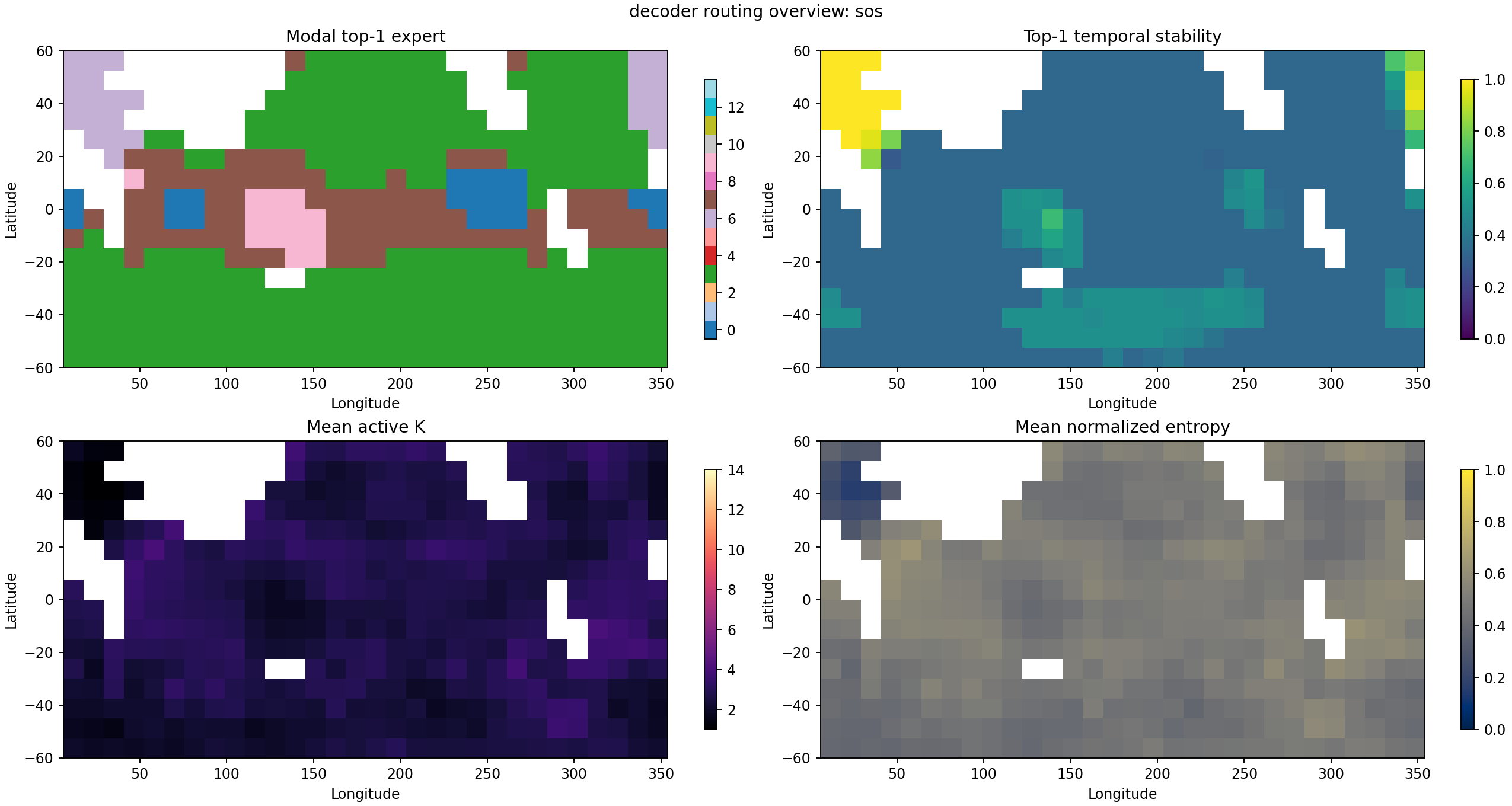}
    \caption{Representative decoder routing diagnostics for sea-surface
    salinity under AdaLN+Geo-12000. The four panels show modal top-1 expert,
    temporal top-1 stability, mean active expert count, and normalized routing
    entropy on the native $16\times30$ grid. White cells have zero ocean
    fraction and are excluded from routing statistics.}
    \label{fig:routing_overview}
\end{figure}

\noindent\textbf{Q8: Does the geographic router organize allocation across
space?}\\[0.25em]
\textbf{A8.} The decoder routing distributions have spatial Jensen--Shannon
divergence values of 0.0318 and 0.0451 for the two GeoRouter configurations,
and their spherical-harmonic fits reach weighted $R^2$ values of 0.935 and
0.934. These statistics indicate large-scale spatial organization in the
average routing patterns. Figure~\ref{fig:routing_overview} provides a
representative map-level view of this organization for sea-surface salinity.

\section{Conclusion}
\label{sec:conclusion}

Taken together, the experiments support selective rather than uniform
parameter sharing for multivariate ocean forecasting. Under the fixed training
and rollout protocol, OceanMoE reduces aggregate error under both evaluated
output contracts and maintains lower geometric-mean normalized RMSE than the corresponding
baselines over most later rollout months. The fusion-weight analysis shows
that target states use distinct mixtures of source variables, while routing
diagnostics show that expert allocation varies across targets and locations.

The target- and region-wise results further show that the measured benefits
depend on the output contract and geographic context. Full-depth sensitivity
tests confirm that the complete architecture has the lowest geometric-mean
ten-variable error ratio, with the largest degradation under fixed top-2
routing. Together, these results support structured conditional computation
as a way to combine shared ocean context with target- and location-dependent
specialization.

{\fontsize{8pt}{8.2pt}\selectfont
\setlength{\bibsep}{0pt}
\bibliographystyle{unsrtnat}
\bibliography{references}
}

\end{document}